\documentclass{article}
\usepackage{natbib}
\setcitestyle{numbers,square}
\usepackage{arxiv}
\usepackage[utf8]{inputenc} % allow utf-8 input
\usepackage[T1]{fontenc}    % use 8-bit T1 fonts
\usepackage{hyperref}       % hyperlinks
\usepackage{url}            % simple URL typesetting
\usepackage{booktabs}       % professional-quality tables
\usepackage{amsfonts}       % blackboard math symbols
\usepackage{nicefrac}       % compact symbols for 1/2, etc.
\usepackage{microtype}      % microtypography
\usepackage{lipsum}		% Can be removed after putting your text content
\usepackage{graphicx}
\usepackage{natbib}
\usepackage{doi}

\title{YOLO Series Target Detection Algorithms for Underwater Environments}

\author{{\hspace{1mm}Chenjie Zhang}\\
	Ocean College\\
	Zhejiang University\\
	Zhoushan, China \\
	\texttt{c.zhang@zju.edu.cn} \\
	\And
	{Pengcheng Jiao}\thanks{Corresponding author.}  \\
	Ocean College\\
	Zhejiang University\\
	Zhoushan, China \\
	\texttt{pjiao@zju.edu.cn} \\
}

\renewcommand{\shorttitle}{YOLO Series Target Detection Algorithms for Underwater Environments}

\hypersetup{
pdftitle={A template for the arxiv style},
pdfsubject={q-bio.NC, q-bio.QM},
pdfauthor={David S.~Hippocampus, Elias D.~Striatum},
pdfkeywords={First keyword, Second keyword, More},
}

\begin{document}
\maketitle

\begin{abstract}
You Only Look Once (YOLO) algorithm is a representative target detection algorithm emerging in 2016, which is known for its balance of computing speed and accuracy, and now plays an important role in various fields of human production and life. However, there are still many limitations in the application of YOLO algorithm in underwater environments due to problems such as dim light and turbid water. With limited land area resources, the ocean must have great potential for future human development. In this paper, starting from the actual needs of marine engineering applications, taking underwater structural health monitoring (SHM) and underwater biological detection as examples, we propose improved methods for the application of underwater YOLO algorithms, and point out the problems that still exist.
	% \lipsum[1]
\end{abstract}

% keywords can be removed
%\keywords{YOLO \and Underwater \and Target detection}

\section{Introduction}
With the rise of technologies such as autonomous driving~\cite{li2022cross}, target detection algorithms have been widely and maturely used in terrestrial environments. However, target detection in underwater environments is an issue that is rarely addressed. With the gradual scarcity of land resources, mankind is gradually focusing on the ocean. Underwater target detection algorithms in engineering~\cite{xu2023mad}, ecological research~\cite{zhong2022real}, marine security and defense~\cite{yan2022ship} and other fields have great potential for development. Since the light conditions are completely different from those on land, target detection algorithms for underwater environments face many problems such as poor image clarity~\cite{shen2022biomimetic} and blurring of the shot~\cite{cheng2022detection}, we hope to give an improved solution from the perspective of basic research from both the hardware and software sides at the same time. Hardware resources are scarce underwater, so we selected the YOLO series of algorithms, which do not require much computing power~\cite{zeng2023lightweight}, for our study. In this paper, taking YOLO-UC and YOLO-UH as examples, we propose YOLO application examples for underwater environments, and analyze the problems that still exist.

First is the YOLO algorithm for structural health monitoring of marine structures. Concrete structures have been extensively used in marine and offshore (MO) engineering due to their low bulk density and inherent durability~\cite{hosseinzadeh2022concrete}. Therefore, our goal is to locate the damage identification of underwater concrete. At present, there have been a lot of studies on the identification of concrete damage on land~\cite{cui2022deep}, but for underwater environment, the types of corrosion will be more complex~\cite{pirie2021image}, and the damage types are different from those on land~\cite{jiao2023vision}. In addition, the underwater robot used for shooting will shake under the impact of waves and currents, which will greatly affect the quality of the image. All these factors work against the effect of underwater concrete identification.

Next is the YOLO algorithm for underwater toxic organism detection. Diving is becoming a popular sport as man's desire to explore continues to grow. However, there are a great variety of underwater creatures~\cite{mora2011many} and most of them are not well understood. This on the one hand makes it difficult for people to find poisonous underwater creatures~\cite{engler2012complex} that pose a danger in the process of diving, and on the other hand makes the process of diving a lot less fun. Although existing underwater biometric algorithms have achieved good results in recognition accuracy~\cite{shankar2023comparing}~\cite{sun2022underwater}~\cite{shen2023multi}, all experiments have been conducted based on datasets, and the lack of experimental tests in the field has led to the neglect of an important issue, which is the actual effective detection distance.

In this study, starting from the actual needs of marine engineering applications, taking underwater structural health monitoring (SHM) and underwater biological detection as examples, we propose improved methods for the application of underwater YOLO algorithms, and point out the problems that still exist.

\section{METHODS}
\label{sec:headings}
\subsection{YOLO-UC}

Firstly, the YOLO-UC (Underwater Concrete) model for health monitoring of underwater structures is introduced. Concrete is the most basic material that constitutes a building, and it is crucial to monitor the structural health of concrete. Especially when concrete is immersed in seawater for a long period of time, which is more susceptible to spalling, corrosion and other damages, it will have an impact on the safety of the structure. At present, robotic inspection is gradually replacing the time-consuming and laborious manual inspection, but there is still a lack of strong technical support for the target detection of underwater concrete damage. Therefore, we propose the YOLO-UC algorithm to solve the practical problems of underwater concrete damage detection based on fundamental research.

\subsubsection{Model preprocessing}
\begin{figure}
    \centering
    \includegraphics[width=1\linewidth]{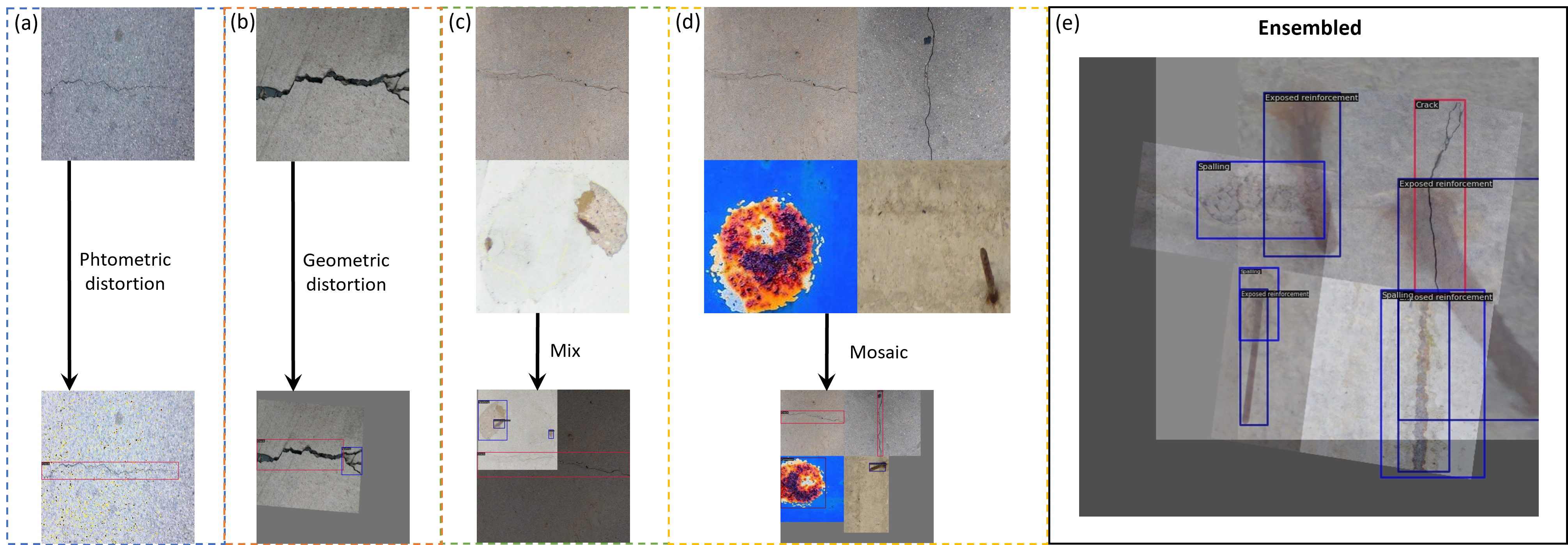}
    \caption{Data augmentation techniques.(a) Photometric distortion.(b) Geometric distortion.(c) Mix.(d) Mosaic.and (e) Ensembled results.}
    \label{fig:1}
\end{figure}

Data augmentation is a commonly used method to boost the performance of a detector and reduce the risk of overfitting. Figure 1 shows four types of data augmentation techniques applied in the training pipeline, namely photometric distortion, geometric distortion, mix~\cite{zhang2017mixup} and mosaic. Firstly, photometric distortion is adopted to reduce the model’s sensitivity to changes of illumination conditions in different underwater scenes. Secondly, due to the changing positional relationship between the camera system and the damaged surface in practical applications, the perspective and scale of captured images can vary greatly. Geometric distortion is used to simulate this effect. Finally, mix augmentation algorithm and mosaic augmentation algorithm, which have been proved effective in YOLOv4~\cite{bochkovskiy2020yolov4} and YOLOv5’s~\cite{redmon2016you} implementation, are leveraged to improve the model's capability of detecting objects in complex backgrounds. Experiments have shown that this data augmentation strategy improved the model’s performance significantly on this built dataset.
To improve the convergence speed of the training loss and achieve more accurate prediction, we introduced Generalized IoU (GIoU) Loss~\cite{rezatofighi2019generalized} to the model to replace the original Intersection over Union (IoU) loss. In YOLOX, bounding box regression is used to predict target objects' location using rectangular bounding boxes. It aims to refine the location of a predicted bounding box. Bounding box regression uses overlap area between the predicted bounding box and the ground truth bounding box referred to as IoU loss. IoU loss, however, only works when the predicted bounding boxes overlap with the ground truth box. It would not provide any moving gradient for non-overlapping cases. Instead, GIoU loss maximizes the overlap area of the ground truth and predicted bounding box. It increases the predicted box's size to overlap with the target box by moving slowly towards the target box for non-overlapping cases. With the introduced GIoU loss function, the test performance of the model witnessed a considerable enhancement.
Three sets of comparison experiments were conducted: the plain YOLOX model, the YOLOX model using the data augmentation strategy (YOLOX-D), and the YOLOX model using the data augmentation strategy as well as the GIoU loss function (YOLOX-DG). Figure 2(a-b) shows the change in mAP in the three experiments. After 80 epochs of training, the mAP curves tend to smooth out and overfitting of the model appears, so we only show the results of 80 rounds of training, which already represents the upper limit of the model. The results shows that the YOLOX model with data augmentation strategy (YOLOX-D), though slower in convergence speed, has a better convergence result compared with the plain YOLOX model. The final mAP of the YOLOX-D model is 57.0 for mAP0.5:0.95 and 76.0 for mAP0.5, significantly better than that of the plain YOLOX model (56.0 for mAP0.5:0.95 and 73.7 for mAP0.5). Moreover, the YOLOX-DG model outperforms the YOLOX-D model, reaching 58.5 for mAP0.5:0.95 and 78.0 for mAP0.5.

\begin{figure}
    \centering
    \includegraphics[width=1\linewidth]{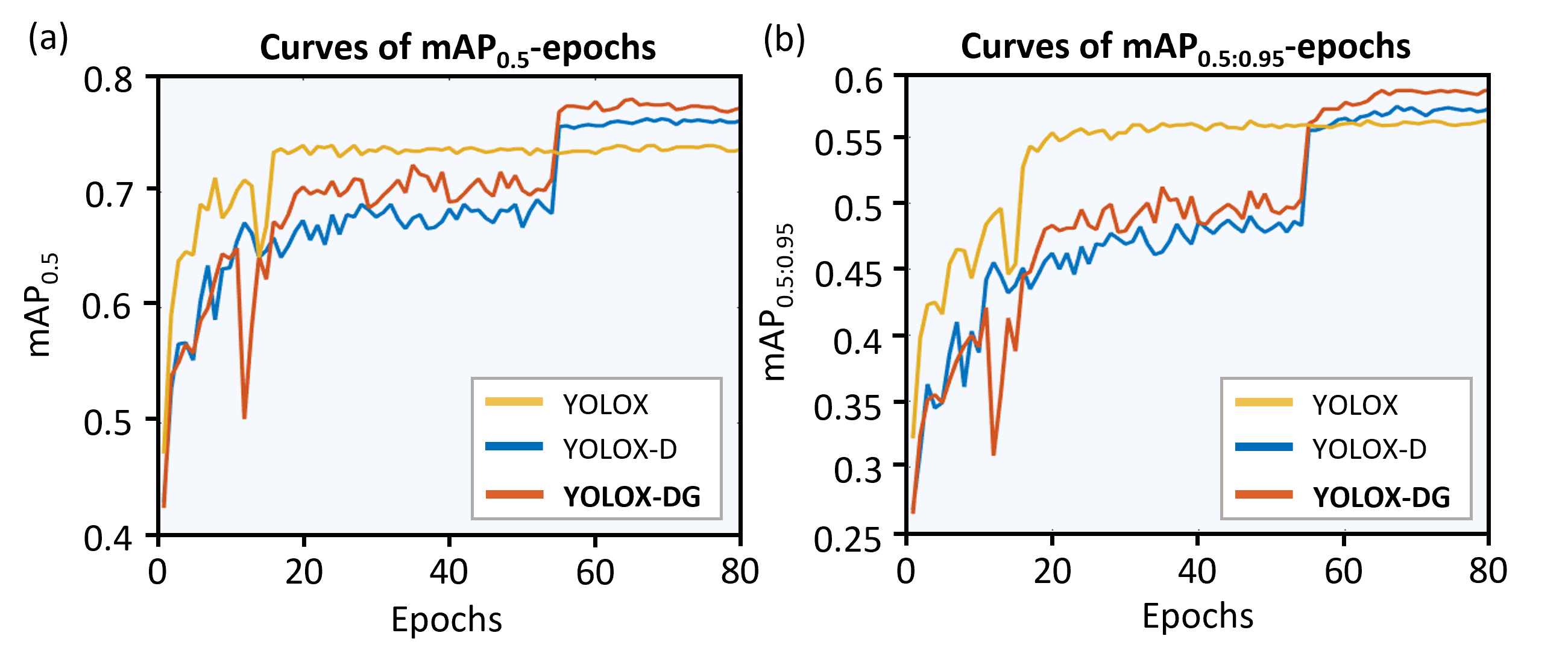}
    \caption{Training results. (a) Change in mAP0.5 in three experiments. (b) Change in mAP0.5:0.95 in three experiments.}
    \label{fig:2}
\end{figure}

\subsubsection{Model Application}
We develop the YOLO-UC model based on YOLOX algorithm for the specific application of underwater concrete structures as an industrial application. To accurately analyze these underwater concrete structures, while addressing the relatively low quality and small quantity of the damage images, we mix the land images with underwater images to jointly conduct the model training. We first preprocess the underwater images, convert them into the forms of land images, and then input into the detection function. Comparing with the existing YOLO algorithms, Figure 3 displays the main improvements of the reported YOLO-UC for the specific underwater concrete structures, which includes the steps of transfer learning, underwater modification and early warning.

\begin{figure}
    \centering
    \includegraphics[width=0.75\linewidth]{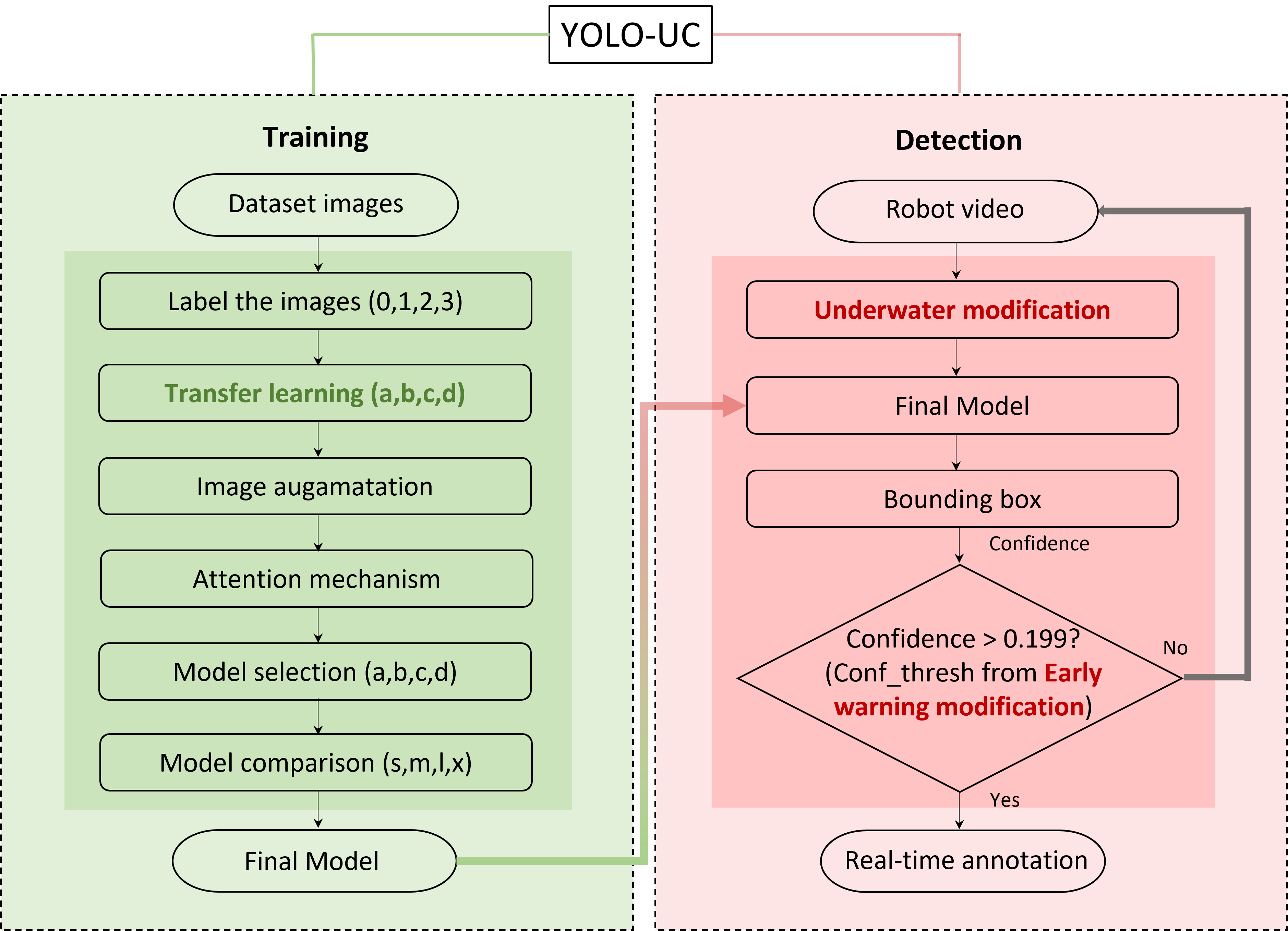}
    \caption{Flowchart of the YOLO-UC model (0, 1, 2, and 3 refer to the four damages of crack, spalling, exposed reinforcement and corrosion, respectively. a, b, c and d refer to the four pre-training models. s, m, l and x refer to the four models of different size).}
    \label{fig:3}
\end{figure}
Transfer learning has been developed to reduce the search scope of possible models in a favorable way by using a model suitable for different related tasks, such that to save computational cost and improve prediction accuracy. In the YOLO-UC model, we first train the model for 50 epochs as the baseline without adding pre-training weights, and then use the training results of Railcrack (https://download.csdn.net/download/gaoyu1253401563/11985328/), MSCOCO (https://cocodataset.org/) and NEU-DET~\cite{bao2021triplet} datasets as the pre-training weights to input the training model. The performance of these four models in 50 rounds of training is shown in Figure 4. We find that the curves of the COCO and Railcrack datasets exceed the baseline, which reflects that the identified objects in the Railcrack datasets are similar to the models. However, we can see from the figure that the COCO curves are slightly higher than the Railcrack curves due to the difference between the label quality and the training level. The NEU-DET dataset is the rail surface damage dataset, which is similar to the concrete surface damages in this paper. Therefore, the mAP value of NEU has obtained great advantages in the same number of training rounds. Based on the transfer learning results, the NEU-based transfer learning is used in the YOLO-UC model to accelerate the training and increase the accuracy.
\begin{figure}
    \centering
    \includegraphics[width=0.5\linewidth]{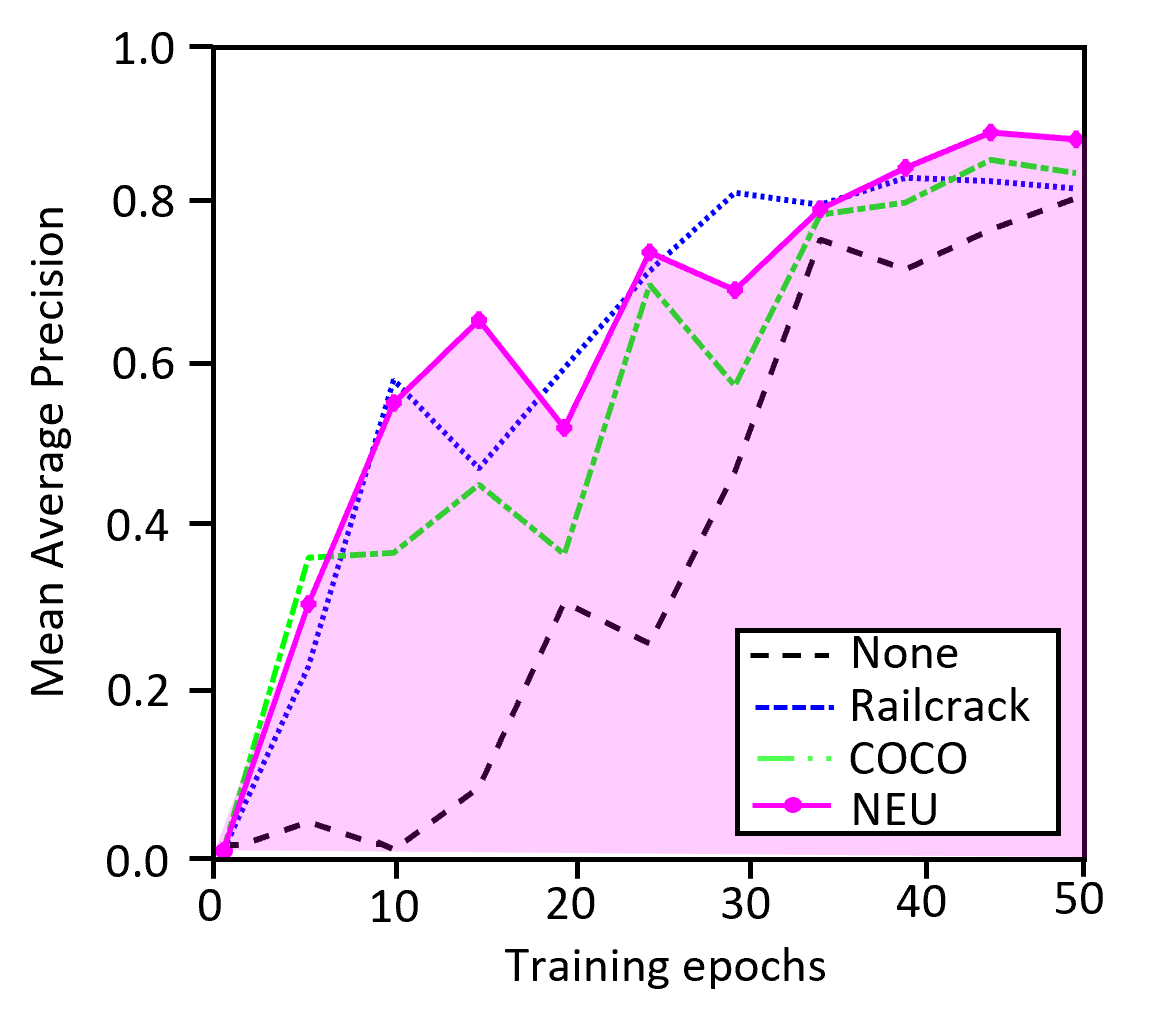}
    \caption{Enhanced performance of the YOLO-UC under the improvements of transfer learning.}
    \label{fig:4}
\end{figure}

In the real underwater environment, the ambient lighting conditions are complex and typically insufficient. Although the underwater robots are designed with light lamp to provide light source, the lighting effect may not be able to fully meet the requirements. In addition, sediment particles floating in underwater severely affect the image clarity of underwater concrete structures. In view of these special problems for underwater images, we preprocess the underwater images, automatically adjust the brightness, brighten dark images, and then appropriately darken the bright images. The Laplace sharpening algorithm is used to reduce the gray level of the central pixel, if the gray level of this central pixel is lower than the average gray level of other pixels in the neighborhood. On the contrary, the gray level of this central pixel is improved if its gray level is higher than the average gray level of other pixels in the neighborhood. The underwater modifications are completely built into the detection function of the YOLO-Underwater model in Pycharm. Figure 5 shows the processing effect of the extremely dark and extremely bright images, as well as the successful recognition results.
\begin{figure}
    \centering
    \includegraphics[width=0.5\linewidth]{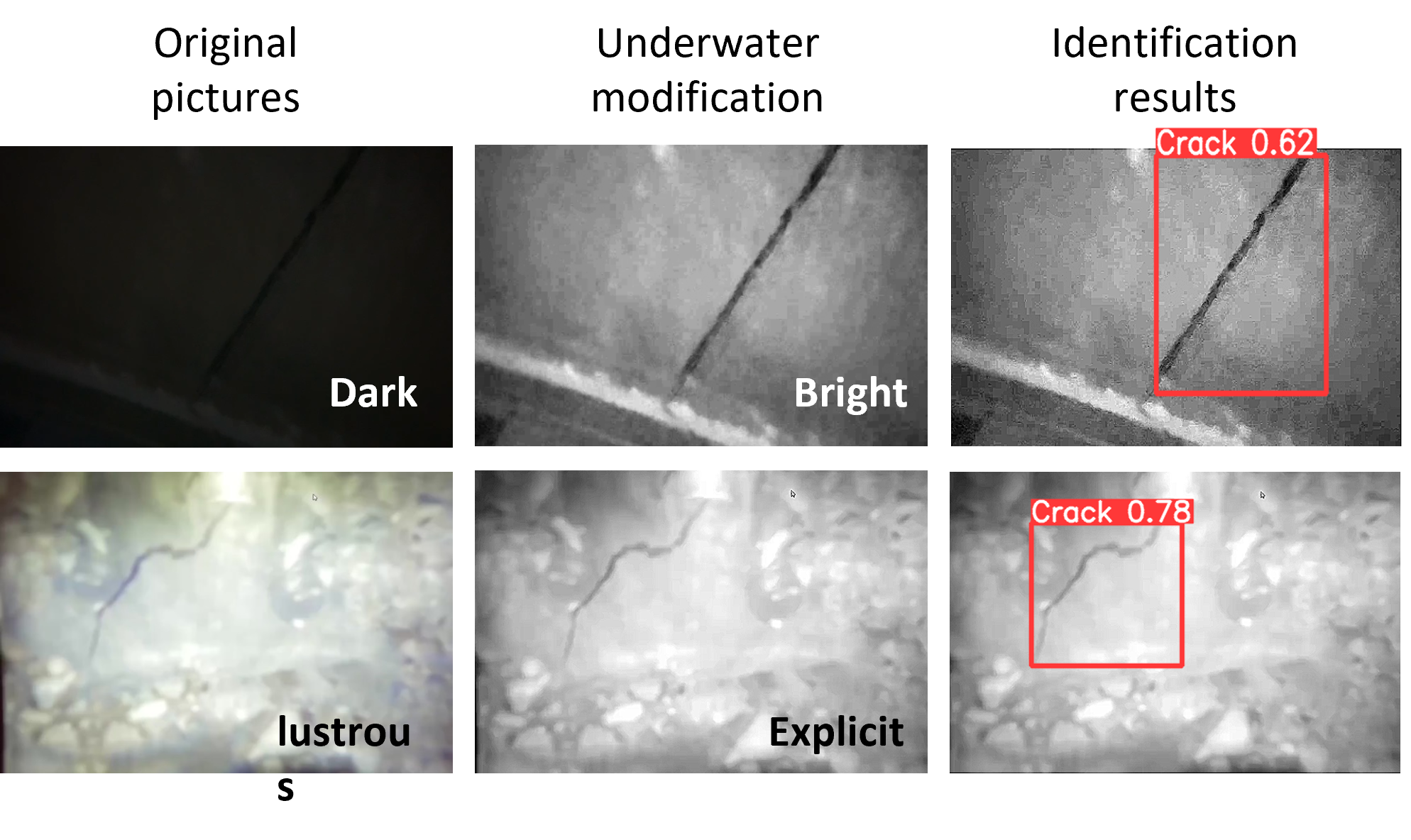}
    \caption{Underwater modification module of YOLO-UC.}
    \label{fig:5}
\end{figure}

The YOLO-UC model is designed with the consideration of damage early warning for underwater concrete structures. Comparing to the situation of missing a potential damage in these concrete structures, the model rather identifying an undamaged area as damage. As a consequence, the YOLO-UC model pays more attention to recall than precision. As shown in Equation (1), the traditional target monitoring models usually set b to be 1 and use F1 score to evaluate the accuracy, that is the harmonic average of the precision and recall. In this model, however, so we set b to be 2 and use F2 score to evaluate the accuracy of the model.
\begin{equation}
F score = (1+b^2)\cdot\frac{Precision \cdot Recall}{(b^2 \cdot Precision)+Recall}
\end{equation}
We set different confidence thresholds to test various types of damage in the model and the general F2 score. Figure 6 shows the curve of the model F2 score under different confidence conditions. When the confidence threshold is set to be 0.199, F2 takes the maximum value of 0.811. Therefore, we set the prediction hyperparameter confidence-threshold as 0.199 in this study.
\begin{figure}
    \centering
    \includegraphics[width=0.5\linewidth]{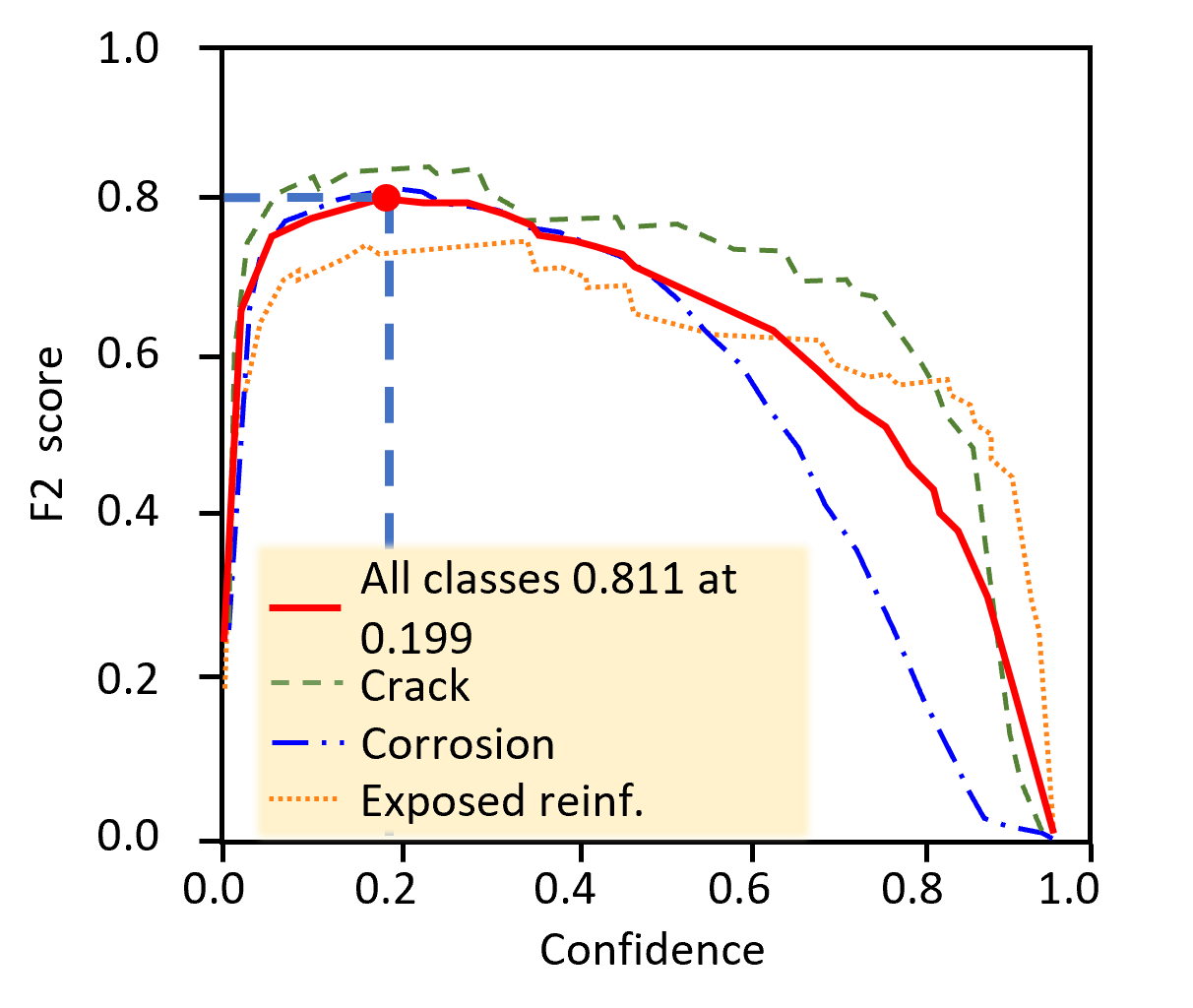}
    \caption{Early warning module of YOLO-UC.}
    \label{fig:6}
\end{figure}

\subsection{YOLO-UH}
In addition to structural health monitoring, underwater wearables are another major area where underwater target detection algorithms are needed. For such tasks, based on the Underwater Smart Glasses (USG) we developed, we propose the YOLO-UH (Underwater Hazard) model based on YOLOv5 modified for the detection of underwater toxic organisms, which can serve as a warning to divers.
\subsubsection{Small target improvement}
Effective detection distance is an important parameter in the task of target detection of underwater toxic organisms, and it loses its significance if the dangerous organism comes to a position very close to a human before the algorithm can detect it. To address the above problem, we introduced a series of strategies for small targets based on YOLOv5.
\begin{figure}
    \centering
    \includegraphics[width=1\linewidth]{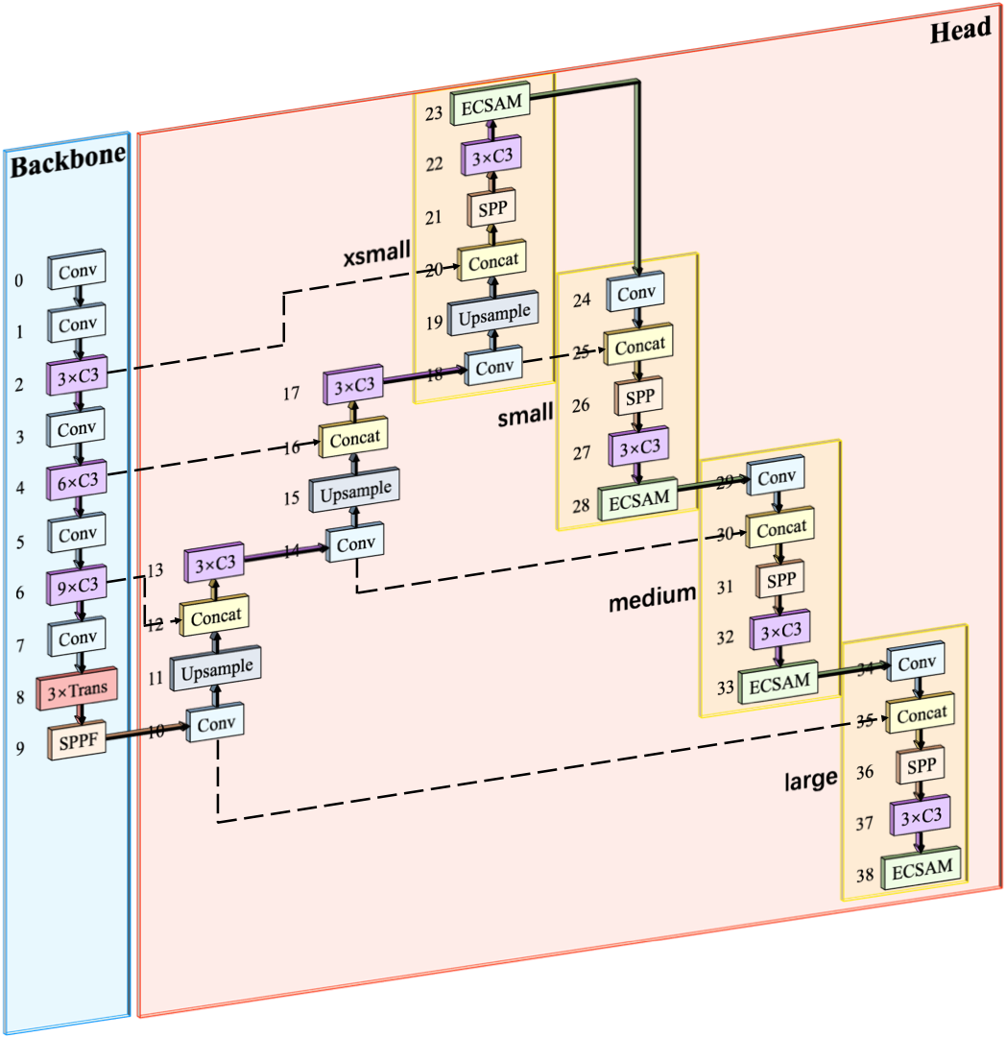}
    \caption{Architecture of YOLO-UH.}
    \label{fig:7}
\end{figure}
Small target detection layer When the size of the input image is 640×640 pixels, detection feature maps of different scales can be obtained by using different feature pyramid layers. Specifically, the P3/8 layer corresponds to a detection feature map size of 80×80 pixels, which is suitable for detecting targets above 8×8 pixels; the P4/16 layer corresponds to a detection feature map size of 40×40 pixels, which is suitable for detecting targets above 16×16 pixels; and the P5/32 layer corresponds to a detection feature map size of 20×20 pixels, which is suitable for detecting targets above 32×32 pixels. It can be seen that the original YOLOv5 detection head does not have a good way to deal with targets below 8×8 pixels, so it can be considered to add a P2/4 small-target detection layer in front of the P3/8 layer, which is used for detecting small targets above 4×4 pixels. In this small target detection layer, smaller receptive fields and higher resolution feature maps can be used to detect small targets. First, three sets of anchor frames need to be generated by k-means clustering based on the self-constructed toxic marine organisms dataset, based on which a set of anchor frames for predicting small targets is added, as shown in Table 1.

\begin{table}
\centering
\begin{tabular}{l}
\hline
anchors: \#4                              \\
-{[}82,96, 134,183, 164,418{]} \# P2/4    \\
-{[}203,256, 277,528, 286,160{]} \# P3/8  \\
-{[}291,336, 378,448, 454,578{]} \# P4/16 \\
-{[}463,307, 529,473, 623,636{]} \# P5/32 \\ \hline
\end{tabular}
\caption{\label{tab:widgets}Small target anchor box settings.}
\end{table}

Next, the feature map needs to be up-sampled to layer P3/8 before further up-sampling and other processes are performed on it to expand the feature map to the size of layer P2/4. At the same time, the feature map with a size of 160×160 pixels is concatenated and fused with the layer 2 feature map P2 in the backbone network to obtain a larger feature map for small target detection. Finally, the small target detection layer is added to the detection layer and a total of four layers are used for detection.

Spatial Pyramid Pooling (SPP) is a feature pooling method for image classification and target detection, which is capable of pooling a feature map of any size of the input image with a fixed size, so that the spatial information of the feature map can be better preserved. The core idea of SPP is to construct a fixed-sized grid, then pool the features in each grid region, and finally stitch all the pooling results together as a representation of that feature map. Compared with the traditional maximum pooling or average pooling methods, SPP can better handle input images of different sizes without losing spatial information, in a way similar to the way we take pictures with different degrees of focus adjustments for different near and far distances in order to clearly capture different distances of the scene. In target detection, SPP is usually used to detect objects at different scales to improve detection accuracy. In this study, SPP was added to the detection head. Here, SPP layers are applied to four feature maps, P2/4, P3/8, P4/16 and P5/32, and each layer uses different window sizes to pool different spatial regions. As a result, targets of different scales can be extracted so that targets of different sizes can be detected efficiently.

Attention Module for Detection Head This study proposes a new attention mechanism, ECSAM (Efficient Channel and Spatial Attention Module), which integrates the advantages of CBAM attention module and ECA attention module. Specifically, the CBAM attention module is more helpful for model accuracy improvement than those attention modules that focus only on the channel dimension due to its spatial attention mechanism; the ECA attention module is relatively more computationally friendly because it performs only one-dimensional convolutional operations without the need to perform global average pooling and fully connected operations. Based on this, the basic principle of the ECSAM attention mechanism is to replace the channel attention in the CBAM attention mechanism with the ECA attention mechanism. In this way, the ECSAM attention module combines the spatial attention mechanism of the CBAM attention module with the one-dimensional convolution operation of the ECA module, which is more suitable for deployment on mobile devices such as underwater smart glasses.

Self-attention mechanism for backbone network The self-attention mechanism, on the other hand, is a special kind of attention mechanism that captures the dependencies between different positions in an input sequence by having each position focus on information from other positions. This mechanism is similar to global attention in human vision, which allows us to see the whole scene instead of focusing only on local details. In the self-attention mechanism, each location has access to information about other locations in the input sequence to better understand the dependencies in the sequence. In this way, the neural network can better utilize the contextual information in the input and improve the accuracy of the model. In this study, Transformer is added to the backbone network as shown in Figure 7. This is because the backbone network is responsible for extracting image features, while the role of Transformer is to optimize the feature representation. By using Transformer, the features extracted by the backbone network will be more accurate and effective, and the detection performance is further improved.

Finally, we compare the improved model with the original YOLOv5 model, and the experiments are conducted on our customized dataset of small targets of underwater toxic organisms. The training results are shown in Figure 8. Both on mAP(0.5) and mAP(0.5:0.95), the improved model with small targets has a higher upper limit of accuracy.

\begin{figure}
    \centering
    \includegraphics[width=0.75\linewidth]{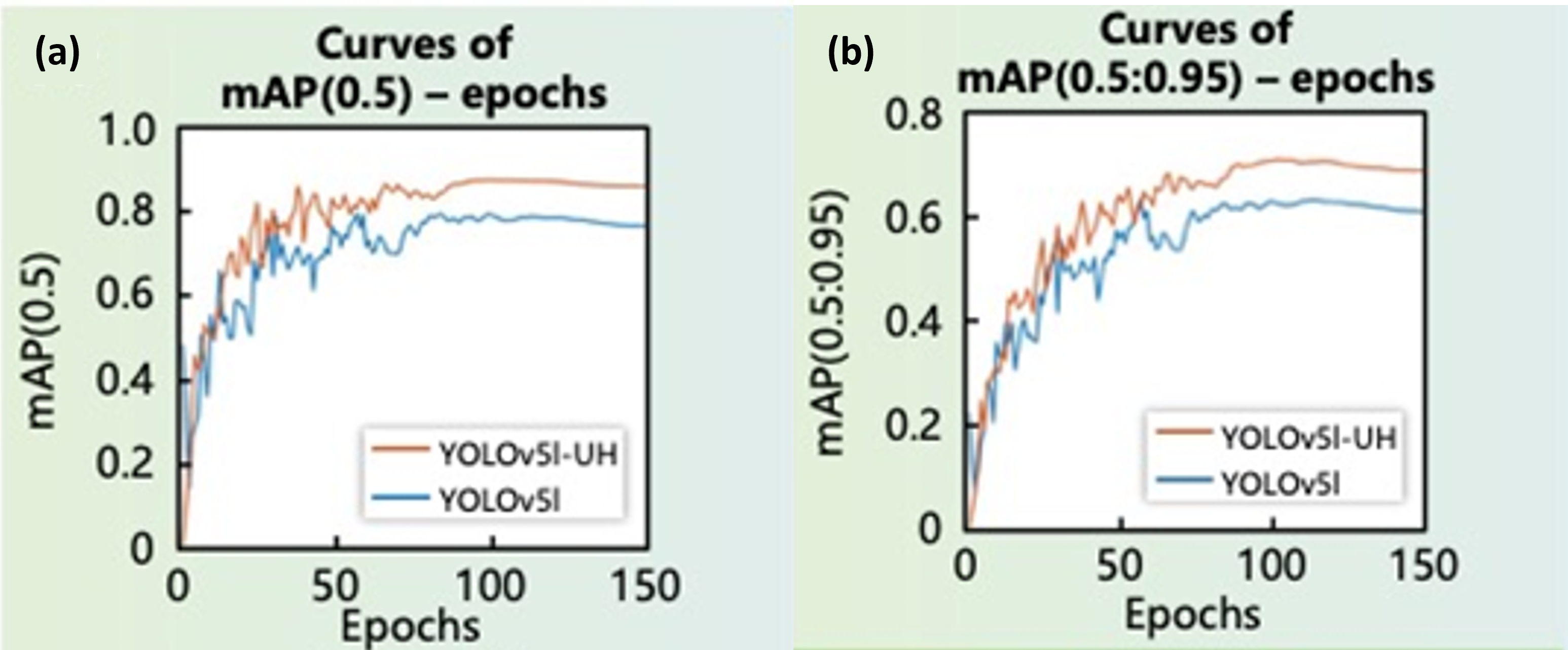}
    \caption{Enter Caption}
    \label{fig:8}
\end{figure}

\subsubsection{Application Analysis}
In this section, we get rid of the limitations of the dataset and conduct experiments directly in real scenarios to observe and analyze the possible problems. For the videos in different situations, we randomly selected 100 frames for recognition, and accuracy is defined as the proportion of them that are recognized correctly. We selected two factors, water turbidity and distance, and conducted an experimental study for four target samples: starfish, sea snake, puffer and conidae.
The experimental results are shown in Figure 9, as the sediment concentration becomes larger, the accuracy first becomes smaller and then larger. It is a common sense situation that the more turbid the water is, the lower the recognition accuracy is. The final trend can be explained by the fact that after the turbidity reaches a certain level, the contrast between the identifier and the background is improved so that the target boundary becomes clearer.
As the distance becomes larger, the accuracy shows a tendency to become larger and then smaller. This may be due to the small number of large targets included in our dataset, resulting in a possible low recognition accuracy in cases where the distance is small but the object boundary can be captured in its entirety. It is worth mentioning that at about 1.5 meters away, the recognition accuracy starts to show a sudden trend of decreasing, and we believe that the performance is unsatisfactory given the small target improvements that have been made. According to our analysis, this happens because the camera is not advanced enough, and there is a focus failure at a distance, resulting in a blurred image.

\begin{figure}
    \centering
    \includegraphics[width=0.75\linewidth]{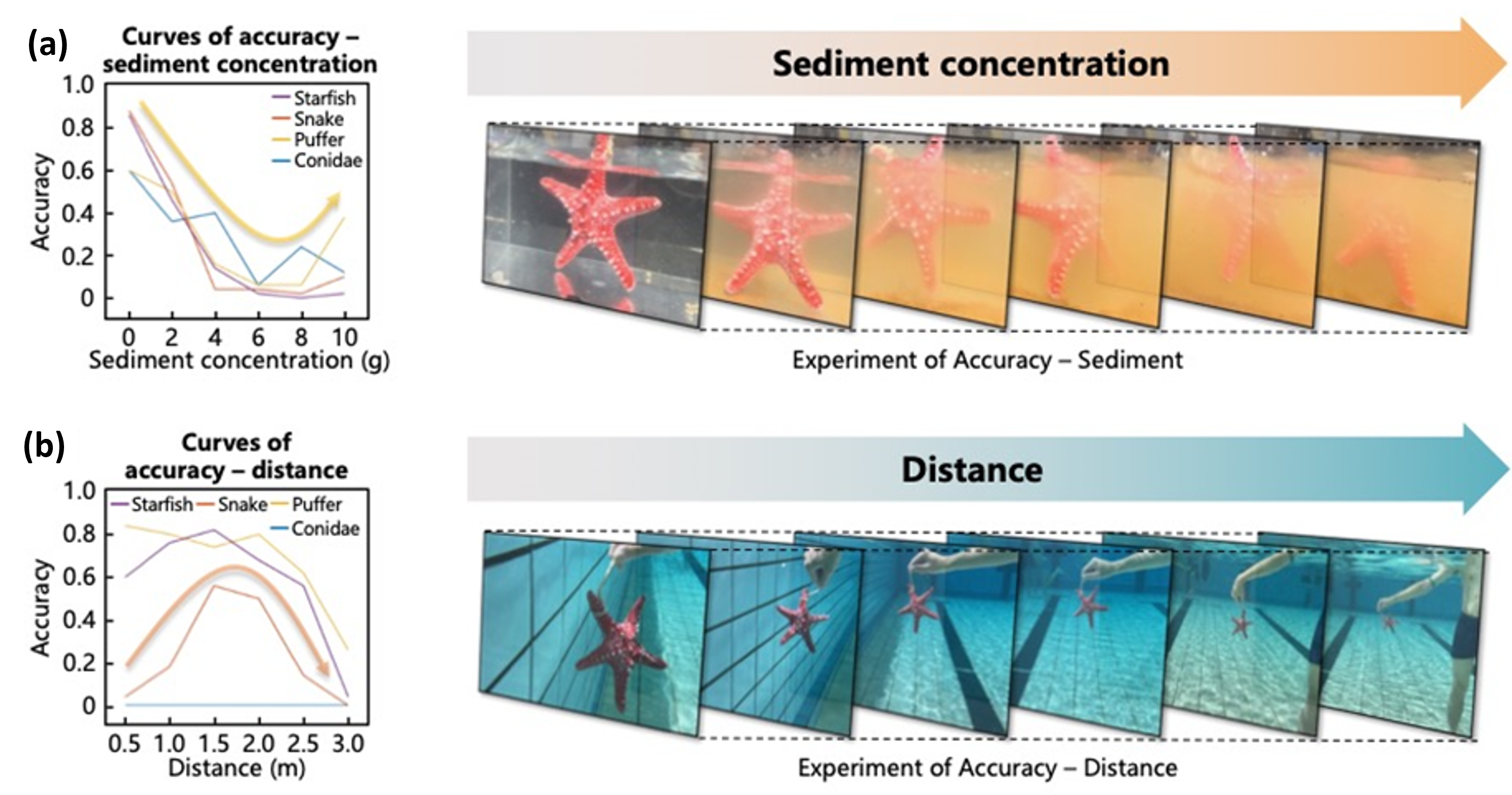}
    \caption{An experimental study of changes in recognition accuracy with sediment concentration and distance.}
    \label{fig:9}
\end{figure}

\section{DICUSSIONS}

The above content shows specific examples of YOLO algorithm in underwater application with YOLO-UC and YOLO-UH as an example. In the process of model development, we have encountered some problems and challenges that still remain to be solved, and we hope to make a sharing and discussion in this section.

As shown in Figure 10, we believe that the problems and challenges faced in YOLO-Underwater+ can be categorized into two aspects: Hardware and Software. First, let's look at the hardware aspect, the first one is the camera problem, the camera's pixel and autofocus functions are especially important in the underwater environment, because human beings can't intervene in it, and all the shooting must be done automatically by the camera. If the quality of the image is very low, then all the rest of the process is out of the question.

The next most important thing is the chip problem, we know that the YOLO model still has some requirements on computing power even as a lightweight target detection model. In the underwater environment, we have to choose a chip with enough power within the limited volume, and at the same time, the heat dissipation also needs to keep up with the supporting hardware, there are a lot of places that can be improved.

The final aspect of the hardware is communication. It is well known that water is a serious obstacle to electromagnetic waves, which makes it difficult for our underwater equipment to interact with the outside world. Currently there are two mainstream solutions, including work with cables and hydroacoustic communications.

\begin{figure}
    \centering
    \includegraphics[width=1\linewidth]{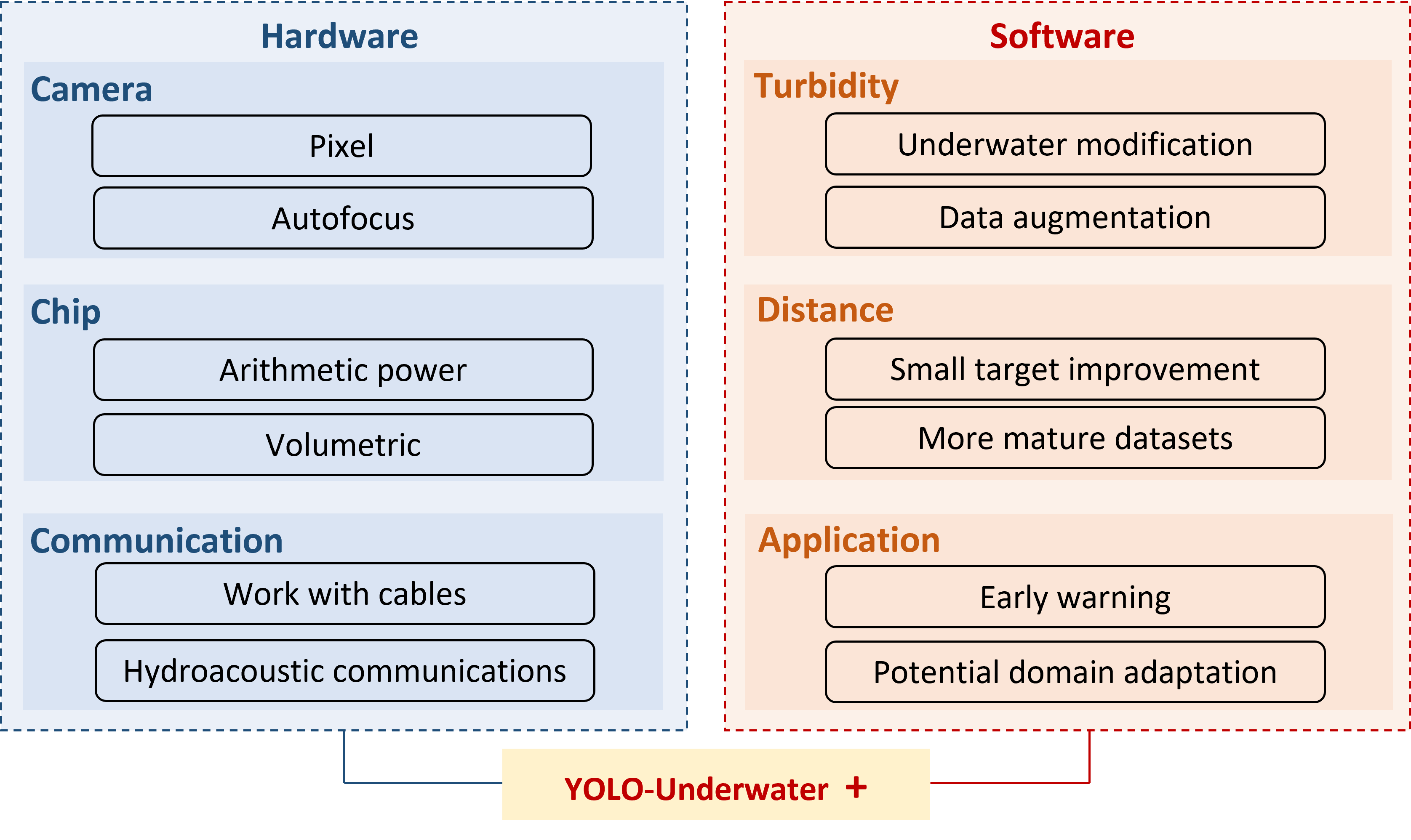}
    \caption{Problems and challenges for YOLO-Underwater+.}
    \label{fig:10}
\end{figure}

The first problem on the software is turbidity, we can deal with this problem in two ways. underwater modification and by means of data preprocessing, for underwater images, preprocessing techniques can be used to improve the image quality, such as de-fogging, contrast enhancement, noise reduction, etc. This helps to improve the clarity and visibility of the image, thus improving the target detection. This helps to improve the clarity and visibility of the image, thus improving the target detection. Data augmentation: improves the robustness of the model by increasing the diversity of underwater image data. More diverse data samples can be generated by data enhancement techniques such as rotation, panning, flipping, scaling, etc.

The next issue is the recognition distance, we hope to improve the model's detection of small targets through stronger small target improvement, and at the same time to build a more mature dataset from the source. At the same time, we hope to build a more mature dataset from the source, and if one day the effective recognition distance can exceed the level of the human eye, it will be very beneficial for human beings to explore the underwater world.

The final aspect of the software is the application side of the problem, we proposed the idea of early warning, to a certain extent, as much as possible, the output of the recognition box to reduce the probability of missed detection, which is conducive to improve the efficiency of the use. However, the early warning model is still not perfect, and the determination of b needs more rigorous calculation and inference. Finally, there is the potential domain adaptation problem, which may exist when migrating the YOLO algorithm from a terrestrial environment to an underwater environment. Due to the differences between underwater and land environments, corresponding domain adaptation methods may be needed to improve the performance and robustness of the algorithm.

\section{CONCLUSION}
This paper focuses on the application of YOLO series algorithms in underwater environments. Aiming at the problem of underwater concrete recognition, this paper adds data augmentation, Laplace sharpening, early warning and other modules on the basis of YOLOX model, and puts forward YOLO-UC model. Aiming at the problem of underwater toxic biometrics, this paper adds a variety of small target optimization modules on the basis of YOLOv5 model, proposes YOLO-UH model, and analyzes the results of field tests. In the Discussion chapter, the existing shortcomings in the field of underwater target detection are analyzed in detail, we will put effort into these shortcomings in our future works!

\bibliographystyle{unsrt}
\bibliography{references}  %%% Uncomment this line and comment out the ``thebibliography'' section below to use the external .bib file (using bibtex) .

%%% Uncomment this section and comment out the \bibliography{references} line above to use inline references.
% \begin{thebibliography}{1}

% 	\bibitem{kour2014real}
% 	George Kour and Raid Saabne.
% 	\newblock Real-time segmentation of on-line handwritten arabic script.
% 	\newblock In {\em Frontiers in Handwriting Recognition (ICFHR), 2014 14th
% 			International Conference on}, pages 417--422. IEEE, 2014.

% 	\bibitem{kour2014fast}
% 	George Kour and Raid Saabne.
% 	\newblock Fast classification of handwritten on-line arabic characters.
% 	\newblock In {\em Soft Computing and Pattern Recognition (SoCPaR), 2014 6th
% 			International Conference of}, pages 312--318. IEEE, 2014.

% 	\bibitem{hadash2018estimate}
% 	Guy Hadash, Einat Kermany, Boaz Carmeli, Ofer Lavi, George Kour, and Alon
% 	Jacovi.
% 	\newblock Estimate and replace: A novel approach to integrating deep neural
% 	networks with existing applications.
% 	\newblock {\em arXiv preprint arXiv:1804.09028}, 2018.

% \end{thebibliography}

\end{document}